\documentclass[conference]{IEEEtran}
\usepackage{cite}
\usepackage{amsmath,amssymb,amsfonts}
\usepackage{algorithmic}
\usepackage{graphicx}
\usepackage{textcomp}
\usepackage{xcolor}

\usepackage{microtype}
\usepackage{subfigure}
\usepackage{booktabs} 
\usepackage{xcolor}
\usepackage{amsmath}
\usepackage{amsfonts}
\usepackage{bm}
\usepackage{algorithm}
\usepackage{algorithmic}
\usepackage{makecell}

\usepackage{hyperref}

\def\BibTeX{{\rm B\kern-.05em{\sc i\kern-.025em b}\kern-.08em
    T\kern-.1667em\lower.7ex\hbox{E}\kern-.125emX}}

\begin{document}

\title{Learning to Generate Levels From Nothing}

\makeatletter
\newcommand{\linebreakand}{%
  \end{@IEEEauthorhalign}
  \hfill\mbox{}\par
  \mbox{}\hfill\begin{@IEEEauthorhalign}
}
\makeatother

\author{\IEEEauthorblockN{Philip Bontrager}
\IEEEauthorblockA{
\textit{TheTake} \\
New York, New York \\
pbontrager@gmail.com}
\and
\IEEEauthorblockN{Julian Togelius}
\IEEEauthorblockA{
\textit{New York University} \\
Brooklyn, New York \\
julian@togelius.com}
}

\IEEEoverridecommandlockouts
\IEEEpubid{\makebox[\columnwidth]{978-1-6654-3886-5/21/\$31.00~\copyright2021 IEEE \hfill} \hspace{\columnsep}\makebox[\columnwidth]{ }}

\maketitle
\IEEEpubidadjcol

\begin{abstract}
Machine learning for procedural content generation has recently become an active area of research. Levels vary in both form and function and are mostly unrelated to each other across games. This has made it difficult to assemble suitably large datasets to bring machine learning to level design in the same way as it's been used for image generation. Here we propose Generative Playing Networks which design levels for itself to play. The algorithm is built in two parts; an agent that learns to play game levels, and a generator that learns the distribution of playable levels. As the agent learns and improves its ability, the space of playable levels, as defined by the agent, grows. The generator targets the agents playability estimates to then update its understanding of what constitutes a playable level. We call this process of learning the distribution of data found through self-discovery with an environment, self-supervised inductive learning. Unlike previous approaches to procedural content generation, Generative Playing Networks are end-to-end differentiable and does not require human-designed examples or domain knowledge. We demonstrate the capability of this framework by training an agent and level generator for a 2D dungeon crawler game.
\end{abstract}


\section{Intro}
\label{intro}

Learning to generate levels is an interesting problem that combines an understanding of aesthetics and style along with the functional requirements of playing a game. Optimization and search based techniques can be very effective level generators when a good objective is defined. This is challenging to get right because many components, especially the aesthetic parts, are difficult to objectively define. Machine learning methods are very good at replicating visual patterns and design patterns but they struggle with a lack of data and often miss the importance of functional components in a level. Machine learning methods can also struggle with discovering new patterns which is often a key use for procedural content systems. What is needed is an approach that learns directly from interacting with the environment and can create its own data. The algorithm interacts with the environment and learns what patterns work and fail directly within the environment. The benefit of such an algorithm is that a designer can design the game rules and mechanics while the algorithm can find interesting levels, or digital realizations of the game, allowing a form of meta-game design.

We here introduce a novel unsupervised Procedural Content Generation (PCG) method that we believe is the first to attempt to address this goal. Unlike other machine learning techniques, it does not require existing content to learn from or any existing player data. Unlike other search and optimization techniques it does not require access to game-playing agents, a forward-model of the game, or a hand-designed fitness function. Our algorithm, called Generative Playing Networks (GPN) only requires a game definition, implemented as a game engine that has a defined state and action space and returns when the game has been won or lost. Based on this, GPN learns a level generator that can generate playable, non-trivial levels.

Generative Playing Networks is setup as a competition between two networks. One represents an agent that is trying to learn to play a game and model the probability of success at every game-state of the given game. This agent interacts with the environment and attempts to learn a policy that gives it a 100\% chance of winning for every game-state. The second network represents a generative function that is trying to learn the distribution of game-states where the current agent has a 50\% chance of winning. This forms a natural curriculum for the agent while also producing level generators at increasing playing difficulties. The system is fully differentiable with the generator learning directly from the agent's own estimates and requires no content as the agent only plays levels designed by the generator. The two form a symbiotic relationship where the agent keeps expanding its definition of what's winnable, allowing the generator to keep challenging that definition.

We demonstrate this approach in a simple 2D dungeon crawling game, which is part of the GVGAI suite of games. However, the general framework could be applied to many other problem classes, such as self-driving cars or environment design.

\section{Related Work}
\label{background}
Procedural Content Generation (PCG) in games is a name for various methods that generate game content, such as maps, quests, characters or textures. In particular, procedural generation of levels of various kinds is a commonly studied research problem. This tracks an interest by the game development community, where many video games include some form of level generation~\cite{shaker2016procedural}.

PCG is also important to AI and in particular reinforcement learning research, because it allows for the generation of an arbitrary number of new environments. One benefit of automatically generating many environments is for testing, as well as encouraging, generalization in reinforcement trained agents; it has been found that in many cases, trained agents overfit to the environment they were trained on~\cite{risi2019procedural}. Several authors have tried using some form of PCG to quantify or increase generalization in reinforcement learning~\cite{cobbe2018quantifying,zhang2018study,justesen2018illuminating}.

Existing approaches to procedural content fall into several classes. Most commercial games rely on hand-coded, constructive approaches~\cite{shaker2016procedural}. A popular approach is to use evolutionary algorithms or other population-based stochastic optimizers to cast PCG as a search problem using a game specific fitness score ~\cite{togelius2011search}. More recently, supervised and self-supervised learning has been applied to PCG. Here, a model is trained on existing game content, and can then be sampled to provide more similar content~\cite{summerville2018procedural}. The obvious downside of such approaches is that they require training data in the form of similar game content. The less obvious downside is that the trained models might produce content that is stylistically similar to what they were trained on, but not functional. To remedy this, one can combine search-based approaches with self-supervised learning; for example, the Latent Variable Evolution technique where a generator is trained on existing content and an evolutionary algorithm is used to find latent variables that make the generator network produce content with desirable characteristics (such as playable levels)~\cite{bontrager2018deepmasterprints,volz2018evolving}. Very recently, the use of reinforcement learning for generating game content has been proposed and demonstrated~\cite{khalifa2020pcgrl}. Just like the search-based methods, PCG via RL requires a quality metric to be used for a reward function, but moves the computation from inference to training time.

Assuming that a PCG method uses some kind of objective or reward function, the question remains what designs to reward. One answer is learnability. By including a learning algorithm in the evaluation function, levels can be optimized for learning potential~\cite{togelius2008experiment}. However, such a procedure is computationally very costly. Another perspective is that of the POET system, which attempts to achieve open-ended learning by keeping a population of environments/levels and searching for new ones that are of appropriate difficulty for the agents~\cite{wang2019poet}. Similar open-ended systems have come out contemporaneous with this work that build a POET-like open ended system with RL agents. In PAIRED, a generative agent builds environments that are beatable by one RL agent and not by a second one, providing a natural curriculum for the playing agents and the generator agent \cite{dennis2020emergent}. In Adversarial Reinforcement Learning PCG, the generator agent builds a level up learning from rewards, task by task, as the playing agent succeeds at each task \cite{gisslen2021adversarial}.

The system described here differs in several ways from the previously proposed systems. It's fully differentiable and the level generator is updated with gradient descent directly based on the performance of the agent allowing it to learn the distribution of winnable levels directly from the agent. This sets it apart from other systems where generators are based on search, and also from the recent PCG via RL paradigm where a non-differentiable reward function is used. Learning directly from the agent network allows the generator to target all possible inputs that achieve the appropriate difficulty. This system is, to the best of our knowledge, the first one that does not need a hand-designed objective or examples to learn to generate real human playable levels.


\section{Generative Playing Networks}
\label{gpn}

\begin{figure*}[ht]
\vskip 0.2in
\begin{center}
\centerline{\includegraphics[width=\textwidth]{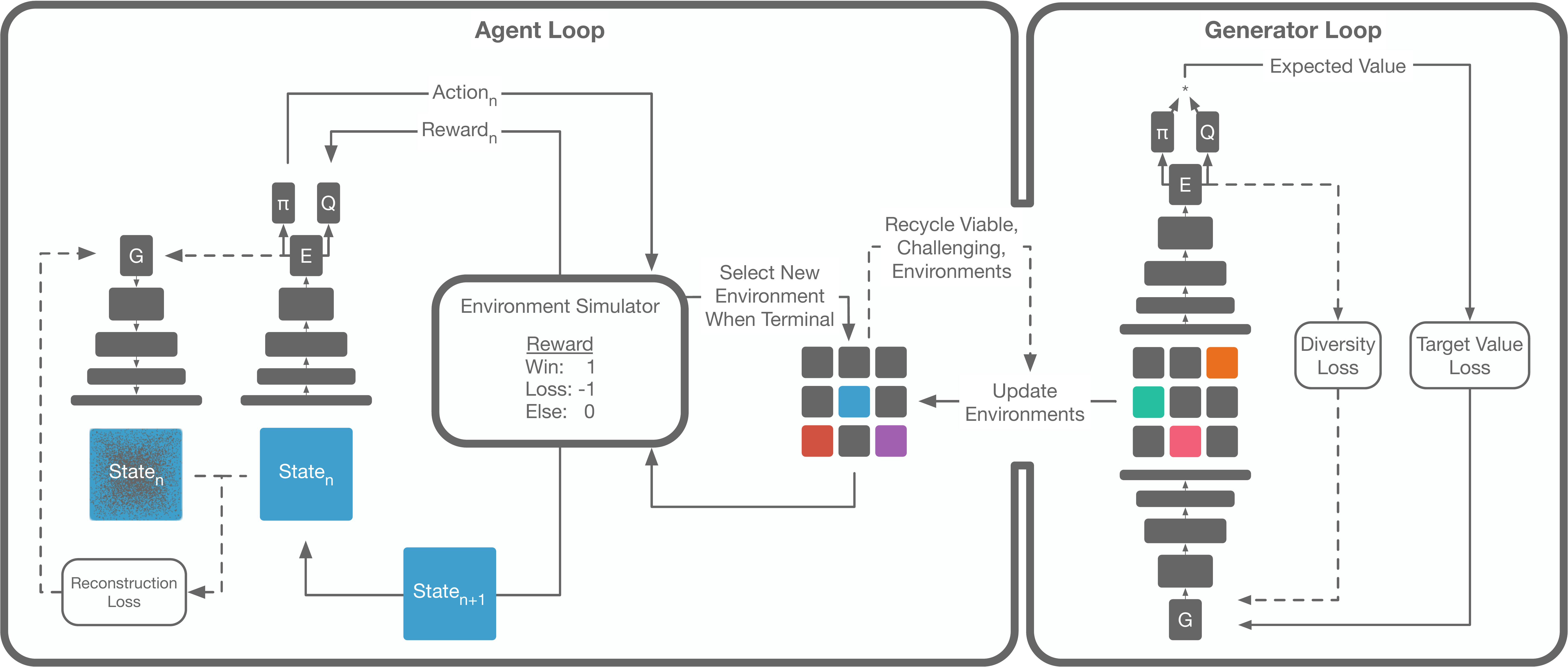}}
\caption{Generative Playing Networks consist of two primary loops; the agent loop and the generator loop. In the agent loop, the agent plays episodes of generated environments. It is learning to estimate the expected total reward from a given state, given the current policy, while also learning an optimal policy for each environment. The generator loop updates the generator to create challenging environments given the agent's current abilities. The dashed lines represent update loops that are not part of the core algorithm but which can help to stabilize training or utilize training data.}
\label{fig:diagram}
\end{center}
\end{figure*}

Generative Playing Networks (GPNs) form a symbiotic relationship between two models, an environment generator $G$, and an agent model. The agent model can further be broken into an environment agent policy $\pi$ and an environment value estimator $Q$. Each model is a differentiable function with parameters $\phi$, $\theta$, and $\omega$, respectively. $G$ is a generator function with the objective of mapping some noise input, $z$, from the distribution $p_z$ to the space of environment states that are reasonably solvable by the agent i.e. environments that the agent has a 5\% chance of winning. The agent is an actor-critic model consisting of both $\pi$ and $Q$. The policy, $\pi$, maps the environment state, $S$, from the distribution of possible environment states for a game, $p_{states}$, to a discrete distribution of the actions available. The state-action utility estimator, $Q$, is trained to estimate the expected value for each possible action given the current state. The expected value for state $S$ can then be calculated as the weighted average of the expected values for each possible action the agent could take, as given in equation \ref{eq:utility}.

\begin{equation}
\label{eq:utility}
    U(\bm{s}) = \sum_{\bm{a \in A}} \pi(\bm{s}, \bm{a}) \times Q(\bm{s}, \bm{a})
\end{equation}
With a definition for calculating the utility of a state, the game between the agent and generator can now be defined. This game is similar in nature to that from Generative Adversarial Networks \cite{goodfellow2014generative}, except the relationship between the agent and generator can be described as both cooperative and adversarial. The agent is attempting to learn a policy that maximizes it’s utility, while the generator is attempting to create environments that challenge the agent. We can thus control the agent and the quality of the generated content through controlling the reward that the agent receives. With a negative total reward for losing and a positive one for winning, the agent will be trying to maximize it’s reward while the generator will learn environments with an estimated utility of 0 as defined in equation \ref{eq:gpn-definition}.
\begin{equation}
\label{eq:gpn-definition}
\begin{split}
\min_G \max_U V(G, U) = & \mathbb{E}_{\bm{s} \sim p_{\text{states}}(\bm{s})}[U(\bm{s})] + \\ & \mathbb{E}_{\bm{z} \sim p_{\bm{z}}(\bm{z})}[U(G(\bm{z}))^2].
\end{split}
\end{equation}
The reason we refer to this relationship as symbiotic, is that the agent and the generator initially cooperate as the agent mostly gets negative rewards in the beginning. The generator will be searching for environment designs that result in higher rewards until it starts generating environments that the agent can succeed in. At this point the agent is receiving positive rewards, and the generator is searching for more challenging environments to lower the total utility, this is where they start competing. The generator is providing a curriculum  for the agent while the agent is providing a complexity measure for the generator.

We then hypothesize that the generated environments will go from, simplistic, to interesting for humans, to unrealistically challenging if the agent is able to gain superhuman performance. Once the algorithm reaches the “interesting for humans” stage, it can be stopped and the generator can be used for sampling many possible reasonable level that can be human curated or it can be searched, via Latent Variable Evolution \cite{bontrager2018deepmasterprints,volz2018evolving} for even more specific environments. An overview of the algorithm is given in figure \ref{fig:diagram} and is defined in full in algorithm \ref{alg:GPN}.

\begin{algorithm}[!htbp]
  \caption{\small Generative Playing Networks (Optional components in blue)}
  \label{alg:GPN}
\begin{algorithmic}
  \STATE {\bfseries Input:} Differentiable policy $\pi_\theta(a|S)$, utility function $Q_\omega(a|S)$, encoder $E_\psi(S)$, and state generator $G_\phi(z)$.
  \STATE {\bfseries Algorithm parameters:} $\alpha^\pi: 2.5e-4$, $\alpha^Q: 2.5e-5$, $\alpha^R: 5e-5$, $\alpha^G: -4$, $m: 128$, $policy updates: 1m$, generator updates: $10$, diversity updates: $90$, pre-training steps: $20m$
  \STATE {\bfseries Extensions:} update model every n steps (n=5), train with small entropy, train with 16 parallel workers, 30\% of elite envs kept, human envs sampled 50\% of episodes
  \STATE Initialize weights \textcolor{blue}{and pre-train agent}
  \REPEAT
    \STATE $z \sim$ minibatch $m$ from $p_z(z)$
    \STATE $S \leftarrow{} G_\phi(z)$
    \STATE Create new environments from generated states
    \STATE \textcolor{blue}{Keep top $k$ levels from previous loop \COMMENT{see \ref{elitism}}} 
    \FOR{number of policy updates}
      \STATE Select environment
      \STATE $S \leftarrow{}$ initialize environment
      \STATE $H \leftarrow{} E_\psi(S)$
      \REPEAT
        \STATE $A \sim \pi_\theta(\cdot|H)$
        \STATE Take action $A$, observe $R$, $S'$ \COMMENT{see equation \ref{eq:rewards}}
        \STATE $H' \leftarrow{} E_\psi(S')$
        \STATE $V' \leftarrow{} \pi_\theta(\cdot|H')\cdot Q_\omega(\cdot|H')$ \COMMENT{If $S'$ is terminal, 0}
        \STATE $\delta \leftarrow{} R + V' - Q_\omega(A|H)$
        \STATE $V \leftarrow{} \pi_\theta(\cdot|H)\cdot Q_\omega(\cdot|H) + \pi_\theta(A|H)\cdot\delta$
        \STATE $Adv \leftarrow{} V' - V$ \COMMENT{If $S'$ is terminal, $R-V$}
        \STATE $\omega, \psi \leftarrow{} \omega, \psi + \alpha^Q \delta^2 \nabla Q_\omega(A|E_\psi(S))$
        \STATE $\theta, \psi \leftarrow{} \theta, \psi + \alpha^\pi Adv \nabla \log \pi_\theta(A|E_\psi(S))$
        \STATE \textcolor{blue}{$Rec \leftarrow{} -\sum_{c} S_c \log G_\phi(E_\psi(S))_c$}
        \STATE \textcolor{blue}{$\phi, \psi \leftarrow{} \phi, \psi + \alpha^R Rec \nabla G_\phi(E_\psi(S))$}
        \STATE $H \leftarrow{} H'$
      \UNTIL{$S'$ is terminal}
      \FOR{number of generator updates}
        \STATE $z \sim$ minibatch $m$ from $p_z(z)$
        \STATE $H \leftarrow{} E_\psi(G_\phi(z))$
        \STATE $V \leftarrow{} \pi_\theta(\cdot|H)\cdot Q_\omega(\cdot|H)$
        \STATE $\phi \leftarrow{} \phi + \alpha^G\ (\frac{V}{m})^2 \nabla_\phi V$
      \ENDFOR
      \textcolor{blue}{
      \FOR{number of diversity updates}
        \STATE $z \sim$ minibatch $m$ from $p_z(z)$
        \STATE $H \leftarrow{} E_\psi(G_\phi(z))$
        \STATE $D \leftarrow{} {(H_a - H_b)^2 \over m}$ for $m$ random pairings of $H_i$
        \STATE $\phi \leftarrow{} \phi + \alpha^G D \nabla_G E_\psi(G_\phi(z))$
      \ENDFOR}
  \ENDFOR
  \UNTIL{User deems environments acceptable}
  \\The gradient-based updates can use any standard gradient-based learning rule. We used Adam in our experiments.
\end{algorithmic}
\end{algorithm}

\subsection{Agent Training Loop}
\subsubsection{Reinforcement Learning Algorithm}
As already discussed, the agent is an actor-critic model that uses reinforcement learning to learn a policy and value function. It is important to have both an actor and a critic in order to properly estimate the expected value of a state. To learn an optimal policy we use advantage actor-critic, as the advantage function has been shown recently to be an effective and stable way to update the actor in many situations \cite{mnih2016asynchronous,justesen2019deep}. The advantage of an action is simply the difference in value between the new state and the old state. The critic is then typically updated incrementally based on temporal difference (TD) learning \cite{sutton2018reinforcement}. Since we’re using the agent to evaluate environments, we cannot simply use the standard critic update.

The agent is used for evaluating environment designs, and as such, it is important that the agent can provide a usable utility for every state of a simulation. In a TD update, the value of state $S_t$ is given as $U(S_t) = R(S_t) + \gamma U(S_{t+1})$ where $\gamma$ is the discount on future rewards and $R(S_t)$ is the reward at $S_t$. If $\gamma < 1$, then the reward will only be able to propagate a short distance backwards in time. In a video game environment with lots of frames, even with a $\gamma$ very close to $1$, the reward is only propagated back a few seconds worth of time. With any discount, very little to no reward from the end of an episode can propagate to the beginning of an episode. Since we need to use our agent to evaluate level designs, and a level is equivalent to the game state at $t = 0$, this is a problem. Thus, for the agent to be able to evaluate environment designs, $\gamma$ must be 1.

This is problematic as it is very difficult to train an agent without discounting future rewards. Without a discount there is very little pressure for the agent to find an optimal path; all possible paths to a reward have the same utility. A random chance of failure for longer solutions help counter this but it's still difficult to learn under these circumstances. We propose to instead explicitly learn the expected value of a state as outlined in equation \ref{eq:utility}.

In actor-critic, the algorithm is trying to maximize the function \cite{sutton2018reinforcement}:

\begin{equation}
\begin{split}
  U^\pi(S) &=  \mathbb{E}[\sum_{n=0}^N \gamma^n R(S_n)]\\
  &=  \sum_{n=0}^N \mathbb{E}[\gamma^n R(S_n)] \\
  &= \sum_{n=0}^N \gamma^n \mathbb{E}[R(S_n)]
\end{split}
\end{equation}

$S_t$ is a random variable and the expectation of the total utility is equivalent to the expectation of the reward at each state. In a typical TD update, the Q function learns this expected value through small updates, basically keeping a running average. This then requires many visits to a state to learn the true expectation of $R(S_t)$ , and many more to learn the true utility when $\gamma = 1$. Since we know the policy distribution $\pi$, we can calculate 

\begin{equation}
\mathbb{E}[R(S_n)] = \sum_{a \in A} \pi(a|S)\cdot \mathbb{E}[R(S_{n+1}^a)].
\end{equation}

Then the expectation of $S_{t+1}^a$ can be represented by Q and the probability of an action is $\pi$.

Algorithm \ref{alg:GPN} outlines the update step as variable $V$. By bootstrapping the expected reward from the policy function, the agent can more easily learn it. In practice, the agent should also be more likely to learn the optimal path. Policy functions are almost never 100\% certain about an action and thus taking multiple steps will naturally discount a future reward from the longer route due to the small uncertainty the policy holds.

While not the main focus of the work, this approach also seems less sensitive to the scale of environment rewards. There is no discount factor that needs to be tuned to match the scale of the rewards to make sure the reward can propagate far enough back in time.

\subsubsection{Agent Reward}
\label{sec:rewards}

For the game, laid out in equation \ref{eq:gpn-definition}, to work, the agent cannot be allowed to learn from the environment's built-in rewards (e.g. game score). Instead the simulator's rewards should be captured and the agent simply given a reward for winning or losing. If a level doesn't compile, the agent instantly loses. In its purest form, this approach is environment independent with 1 point for winning and -1 point for losing. The agent is able to pursue an expected reward of 1 while the generator is able to try and keep the agent's expected reward at 0. 
\begin{equation}
\label{eq:rewards}
 R(S_n)= 
\begin{cases}
    1 & \text{if agent wins}\\
    -1 & \text{if agent loses}\\
    0 & \text{else}
\end{cases}
\end{equation}
If the environment is too difficult to learn in this restricted reward setup, it can be modified to allow more frequent rewards. One modification, that is still domain agnostic, is to scale the reward for winning and losing by two, and then provide a reward of $1\over N$ anytime the environment tries to return a reward. The agent could also be rewarded for surviving longer or winning quickly. These modifications maintain that a positive score is still only achievable by winning and the environment is able to help the agent with more frequent rewards. The generator will also be able to target these rewards as well when learning to make levels with a given expected value. If necessary, the win and loss rewards could be further scaled and custom rewards could be used for a given domain. This is one way that domain knowledge could be added into this approach.

\subsubsection{Environment Selection}
\label{elitism}

During the agent update loop, the agent will play one episode of a environment and then randomly select a new environment from the mini-batch of environments generated. The agent will keep re-sampling from the given minibatch until it has completed a specified number of updates and then the generator will be updated and the minibatch of levels will be replaced with a new set.

To assist with training, two methods can be used with the standard selection; pretraining and elitism. Pretraining is the case where you have access to some example data and can train the networks in a semi-supervised setup. With pretraining, the agent can only select from a curated set of well designed environments until the agent has learned to solve them. This gives the generator a valid direction to learn from at the beginning, otherwise it is essentially generating random environments at the beginning. 

The second assistance comes from an elitism mechanism that keeps the most useful environments around. There are two components to this elitism. The first is to assume that the hand-curated environments are elite and to re-sample them periodically after the pretraining stage. The second is to persist environments that the agent is still learning to play.

After the agent has finished its update loop, the environments can be ranked according to the average reward the agent received. The levels are ranked by the distance to zero of their average score. This is to simulate the generator which is attempting to output levels with an expected reward of zero but this is more stable as the levels are known for sure to have that reward. Once the environments are ranked, the top ones are kept based on a specified percentage of environments to recycle each time.

\subsubsection{State Reconstruction}

To give the generated environments a more human-designed appearance when trained in a semi-supervised setting, the generator can also learn directly from the curated environments. This is done during the agent loop. When the agent encodes the state, this encoding can be passed into the generator, as a latent variable, and the generator can be tasked with reconstructing the input like an autoencoder. For this reason, we actually define the agent as three networks: a policy network, a utility network, and an encoder network that encodes the state and feeds into the first two networks.

This can help constrain the generator to the manifold of possible levels (according to the agent) that look similar to human designed levels. This has been shown to be true with adversarial examples for classifiers, that random noise can cause a network to activate in the same way as the natural images it is trained on \cite{nguyen2015deep}. For this reason, if a classier is used to train a generator, the most likely outcome is a random pattern \cite{nguyen2016synthesizing}. Training the generator to also be a decoder can help mitigate this potential problem. In practice this did not seem to be too big of a problem.

We do not put any constraints on the output of the encoder network, so the generator network is likely learning two separate tasks for two different input distributions. This allows the two tasks to not interfere with each other, while still affecting each other. One could train the decoder as a variational autoencoder to have generating and decoding have the same inputs \cite{kingma2014auto}.

\subsection{Generator Loop}

The generator loop simply consists of updating the generator to create environments with an expected value of 0. The generator is updated by sampling a minibatch of $m$ random latent variables and mapping them to environments which are evaluated by the agent. The generator's weights are then updated. This can be repeated a few times but doing too many updates seems to be detrimental to the generator's diversity.

\subsubsection{Diversity Update}

To help combat the collapse of diversity, a proposed extension for the main algorithm is to update the network a few times explicitly with the goal of increasing diversity. Here, similarity is calculated as the $L2$ difference between the encoding values of two environments. Environments from a minibatch are randomly paired and their similarities calculated. Since every two environments in a minibatch are independently generated, one can simply compare every other sample for similarity. The network is then updated to maximize the average distance between samples. It was found that it is most effective to do the diversity update separate from the primary update target.

\section{Experiments}
\label{experiments}

\begin{figure}[ht]
\begin{center}
\centerline{\includegraphics[width=.9\columnwidth]{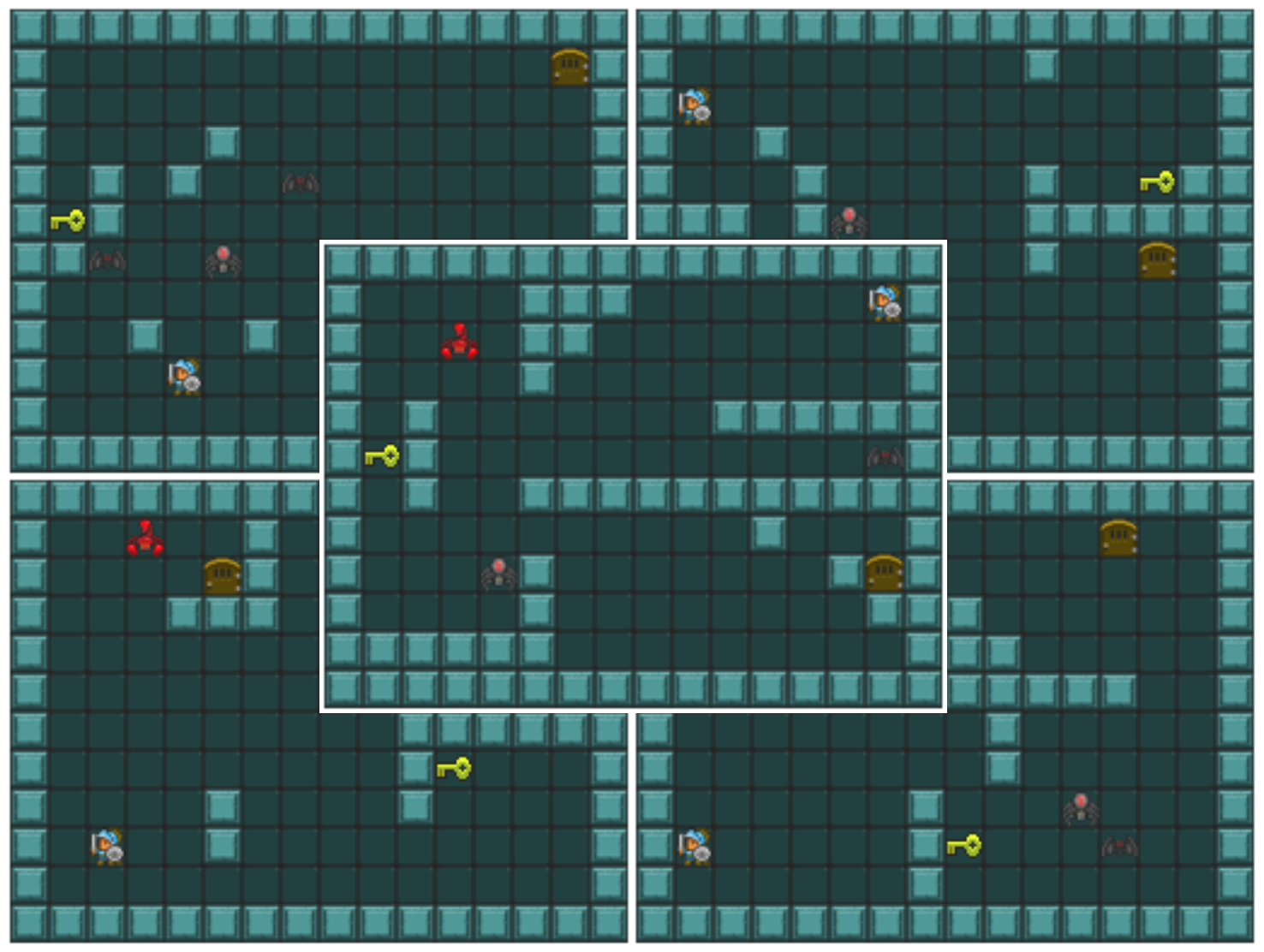}}
\caption{Zelda from the GVGAI framework is a simple dungeon crawler. To beat a level, the player has to collect the key and open the door while avoiding or killing the monsters. Shown above are the five levels provided from the game.}
\label{fig:zelda_original}
\end{center}
\end{figure}

To validate the algorithm proposed here, we test Generative Playing Networks on Zelda from the General Video Game AI (GVGAI) framework \cite{perez2019general}. GVGAI is based on the Video Game Description Language (VGDL) which is a language for describing video games, mostly within the family of arcade games. There is a large corpus of simple games written for this language with each game definition also describing the set of level designs that can exist for it. The framework has been used extensively as a test-bed for game and level PCG research. GVGAI also has an OpenAI gym interface for interfacing  with reinforcement learning algorithms \cite{torrado2018deep,brockman2016openai}.

This is a perfect test-bed for Generative Playing Networks, the agent receives the game-state as a tile representation of the game, where each tile is a one-hot vector of the object in a given location. The generator creates this tile representation state which is directly converted to a level map for the environment to load. If a level does not have an avatar, key, or door, the game engine can't load it and it is presented to the agent as an instant loss after one frame. We run two experiments here, one is the unsupervised setting and the other is semi-supervised using 5 human made levels.

We train Generative Playing Networks on this game and each level with the objective of getting interesting new levels. We do one experiment on the core unsupervised approach, this is the part defined in black in Algorithm \ref{alg:GPN} or the full arrows in Figure \ref{fig:diagram}. We do a second experiment to show how this approach can take advantage of even a few datapoints. For this experiment we use the additional parts of the algorithm: the blue lines in Algorithm \ref{alg:GPN} and dashed arrows in Figure \ref{fig:diagram}. Below we detail some of the experiment parameters. The rest of the parameters we used can be found in algorithm \ref{alg:GPN}.

\subsection{Model Architectures}

The levels being generated for the experiment are 12 by 16 tiles by a one-hot encoding of 14 possible objects. Since the space of legal VGDL levels is smaller than the number of possible game states, we mask out cell types that are not allowed for a level design e.g. an object that must be spawned first. Another approach we could have taken would have been to let the level design compiler automatically select a blank tile when an object is not possible.

Our generator model consists of a fully connected layer that expects a latent vector, of size 512, sampled from a standard normal distribution and transforms it to an output of 3 by 4 by 512. This is processed through two convolutional layers, with kernels of size three, before being passed through a dropout layer and then transformed with a sub-pixel convolution layer to double the output size. This convolution and upsampling is repeated until the expected output size is reached. Each layer contains 512 filters and LeakyReLu is used as the activations. The final output is passed through a Softmax activation to better learn the one-hot encoding. We found that the dropout layer is another helpful tool in maintaining diverse results.

It was found that the level style was fairly susceptible to the up-sampling choice for the architecture of the generator. We got good results with nearest-neighbor upsampling, with transposed convolutions, and with sub-pixel convolutions \cite{radford2015unsupervised,shi2016real}. We chose to do all our experiments with sub-pixel convolutions as it was our favorite aesthetic.

For the agent, we use a nine layer residual network \cite{he2016deep} followed by a gated recurrent unit. The encoded state is a vector of size 512. The policy and utility function are each a single, fully connected layer.



\section{Results}
\label{results}

\begin{figure}[!ht]
\vskip 0.2in
\begin{center}
\centerline{\includegraphics[width=\columnwidth]{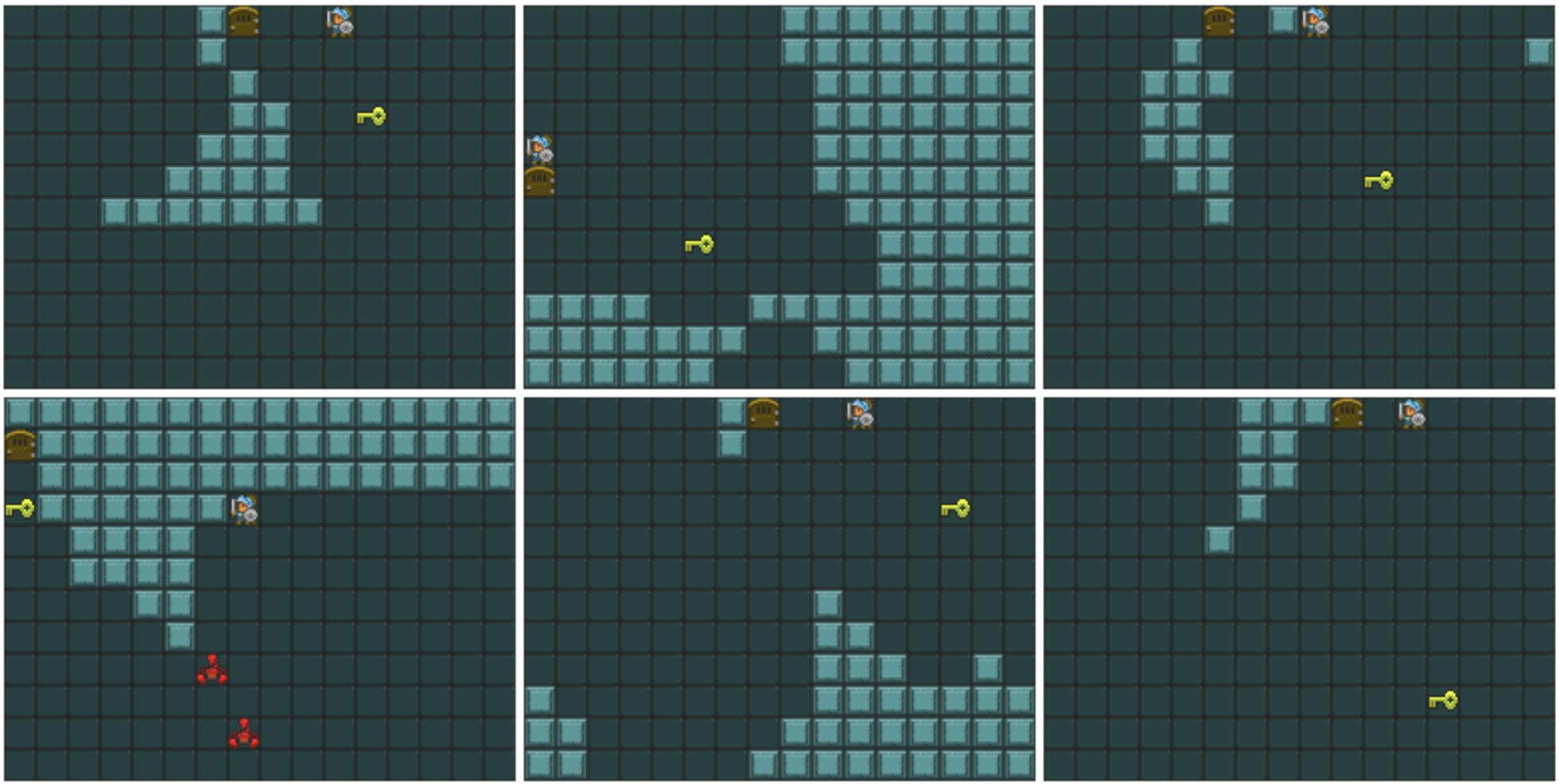}}
\caption{Random Selection of Self-Supervised Generated GVGAI Levels Trained from Nothing}
\label{fig:self_supervised_generated_levels}
\end{center}
\vskip -0.2in
\end{figure}

\begin{figure}[!ht]
\vskip 0.2in
\begin{center}
\centerline{\includegraphics[width=\columnwidth]{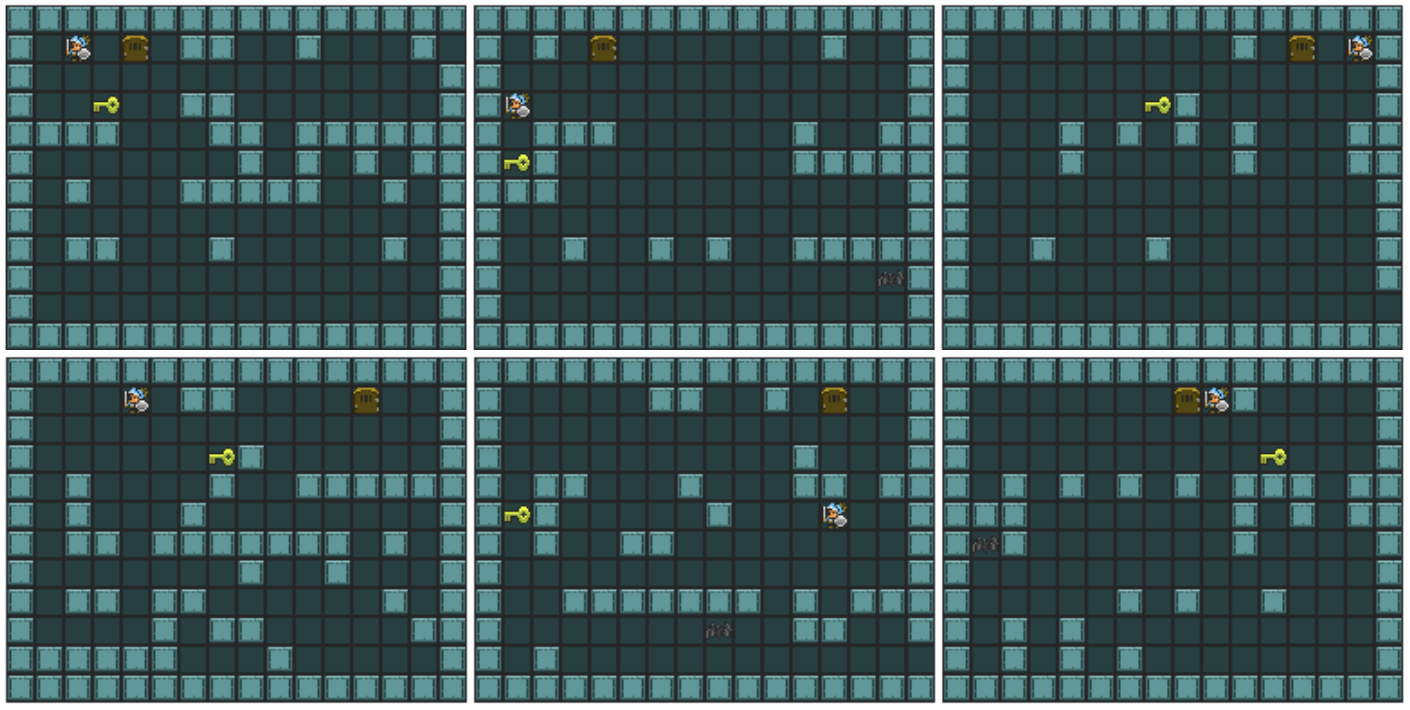}}
\caption{Random Selection of Semi-Supervised Generated GVGAI Levels Trained from 5 Examples}
\label{fig:generated_levels}
\end{center}
\vskip -0.2in
\end{figure}

Figure \ref{fig:generated_levels} and  \ref{fig:self_supervised_generated_levels}  are a random selection of GVGAI Zelda levels generated after training on the game for 50 million frames. After that much training the results tend to collapse into a single design. For both experiments, all of the generated levels are playable/winnable. Only two levels in Figure \ref{fig:generated_levels} and one in Figure \ref{fig:self_supervised_generated_levels} have enemies. 

Training from no examples, the agent seems to learn that open spaces are good for ensuring it doesn't create an impossible level. Outside of one example, it doesn't create obstacles for getting to the door but it does keep the key away from the agent forcing it to find the key for each level. Looking closely at the semi-supervised results, it's clear that the generator learned to emulate the general style of the provided examples but that it's outputs are all unique. All of the playable levels would be considered very simple for a human player. It is understandable that the levels are mostly open, as the agent likely struggled to tell the difference between solvable and unsolvable maze-like environments.


While we were successfully able to train a generator network to create new GVGAI Zelda levels with no example levels and no domain knowledge, we were only able to learn simple levels. This suggests the agent never learned a truly general policy as it could otherwise quickly beat these levels and the generator would have to learn to make more challenging levels to lower the agent's average. This suggest that the algorithm would benefit from benefit from reinforcement learning agents that generalize better across levels.

\subsection{Curriculum}

One of the ways in which the generator and agent work together is that the generator should provide a curriculum for the agent to learn from easy levels and then improve. What our results seem to indicate though, is that the generator first generates complex, unsolvable, designs. Then it generates simple solvable designs and finally very easy designs. This is demonstrated from left to right in figure \ref{fig:generated_curriculum}.
\begin{figure}[H]
\begin{center}
\centerline{\includegraphics[width=\columnwidth]{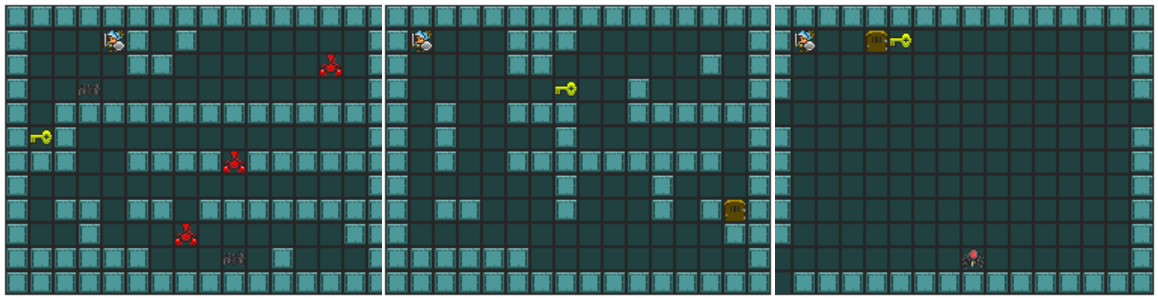}}
\caption{Three sample generated levels chosen from early in the learning process to late in the process to show how the generator learns to match the agent's skill. These were taken from the semi-supervised learning experiment.}
\label{fig:generated_curriculum}
\end{center}
\end{figure}
Figure \ref{fig:generated_rewards} shows that the generator has learned to successfully keep the agent at a reward of 0. Therefore the curriculum we're seeing is from the collaborative half of the algorithm. If the agent is continuously improving, the levels should be getting more difficult according to figure \ref{fig:generated_rewards}. 
\begin{figure}[!htbp]
\begin{center}
\centerline{\includegraphics[width=.9\columnwidth]{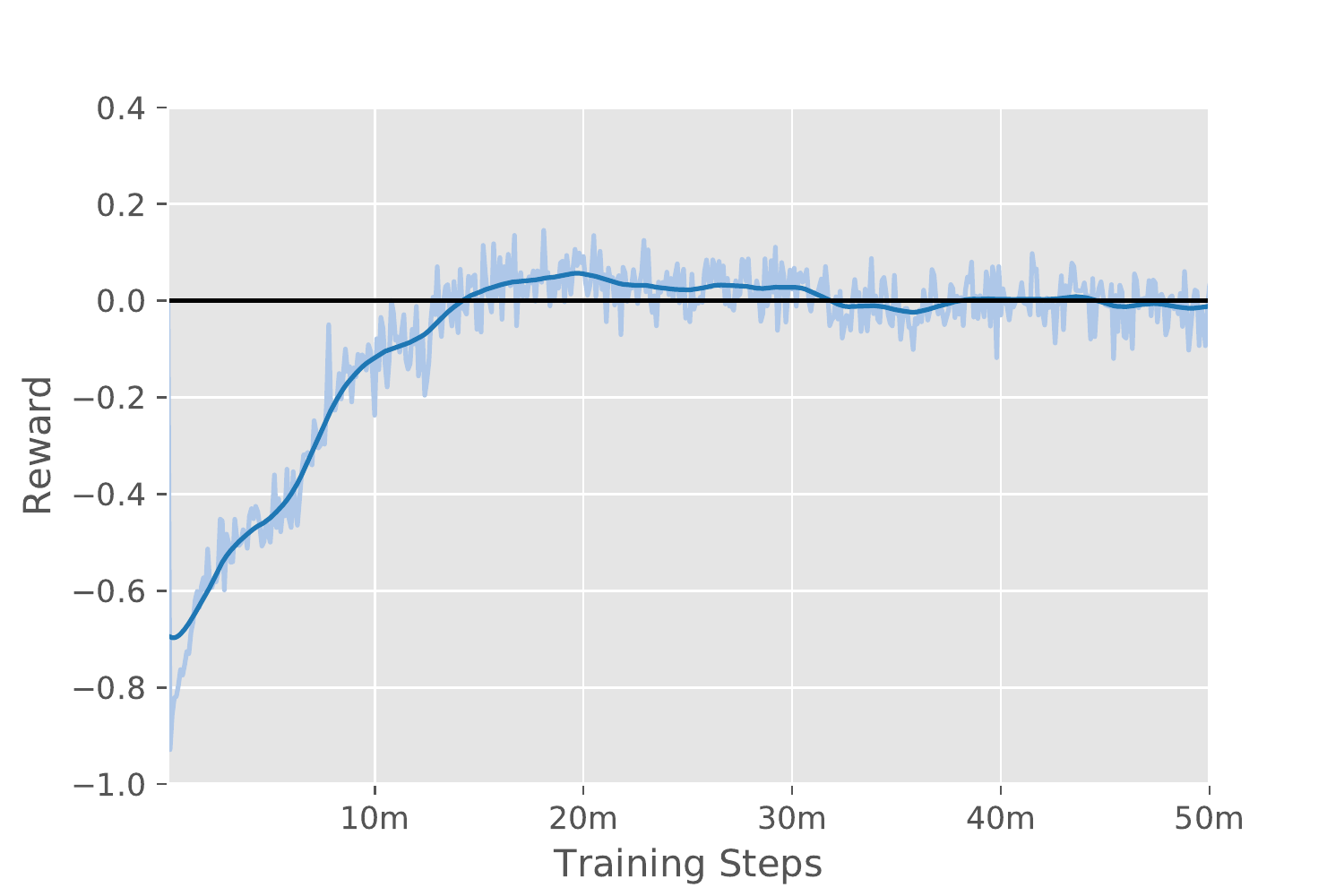}}
\caption{Real average rewards of generated environments converging toward their estimated target.}
\label{fig:generated_rewards}
\end{center}
\end{figure}

\subsection{Agent Results}

We also include here two metrics to show that the agent is learning what it is intended to learn. Figure \ref{fig:est_value} shows a plot of the agent's estimated level value, only from frame 0, versus each level's actual reward at the end of the episode. It's clear that the agent's estimates closely track the agent's actual rewards meaning that it's learning accurately. Though there is much less variance in the estimates than in the real values, this either is reflecting that the estimates are not affected by the stochastic environment or that the estimates are only accurate for the most average levels.
\begin{figure}[!htbp] 
\begin{center}
\centerline{\includegraphics[width=.9\columnwidth]{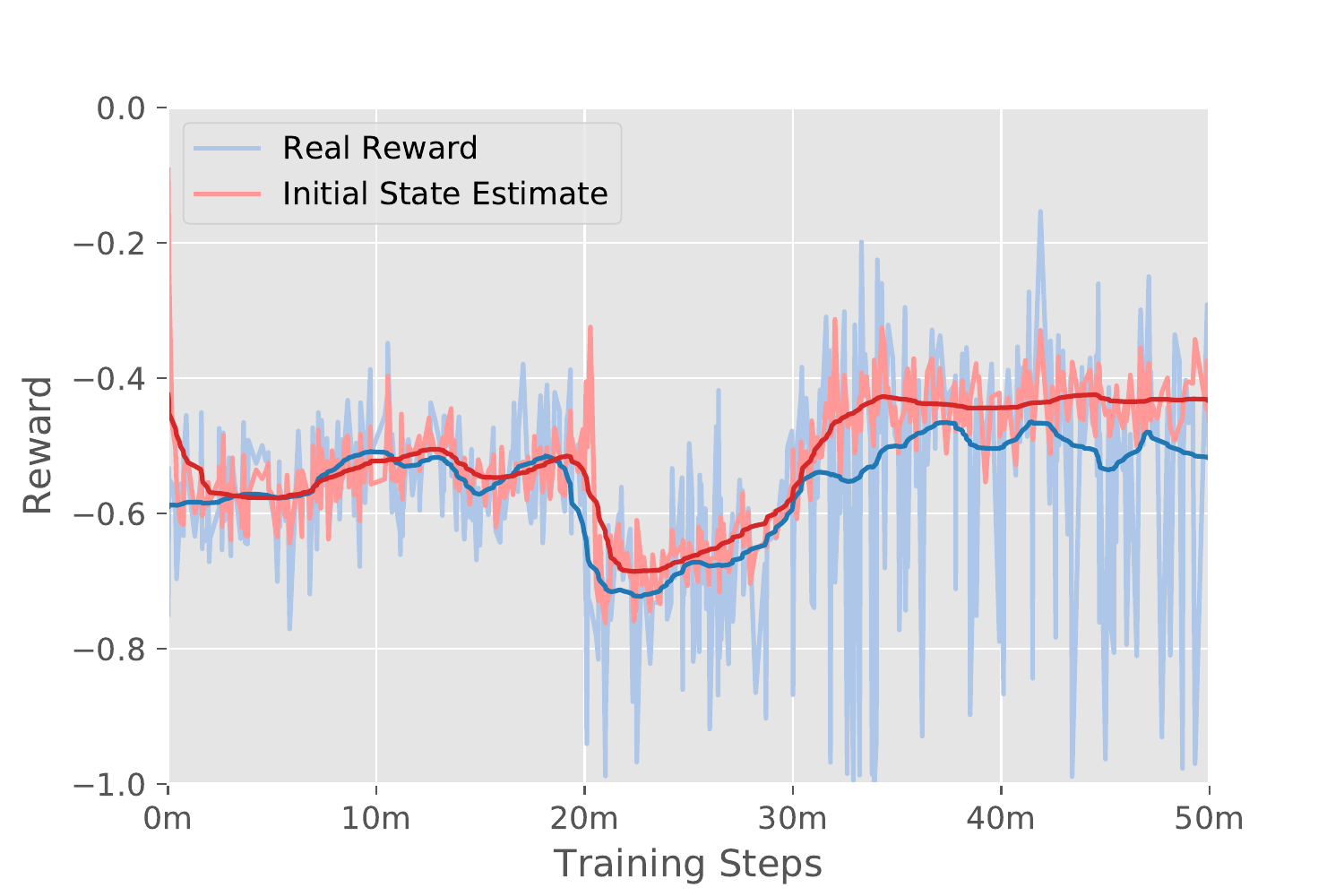}}
\caption{Agent initial state value estimate in red vs real level reward in blue.}
\label{fig:est_value}
\end{center}
\end{figure}
Figure \ref{fig:est_fail} instead shows where the agent's estimates are failing. In the early stages of training, the agent is correctly estimating that impossible-levels have a value of -1, but as it encounters less of them (the red line) it starts to value them higher (the blue line). This either means that later impossible-levels are more exotic and unique, or, more likely, that the agent is forgetting after millions of updates.
\begin{figure}[!htbp]
\begin{center}
\centerline{\includegraphics[width=.9\columnwidth]{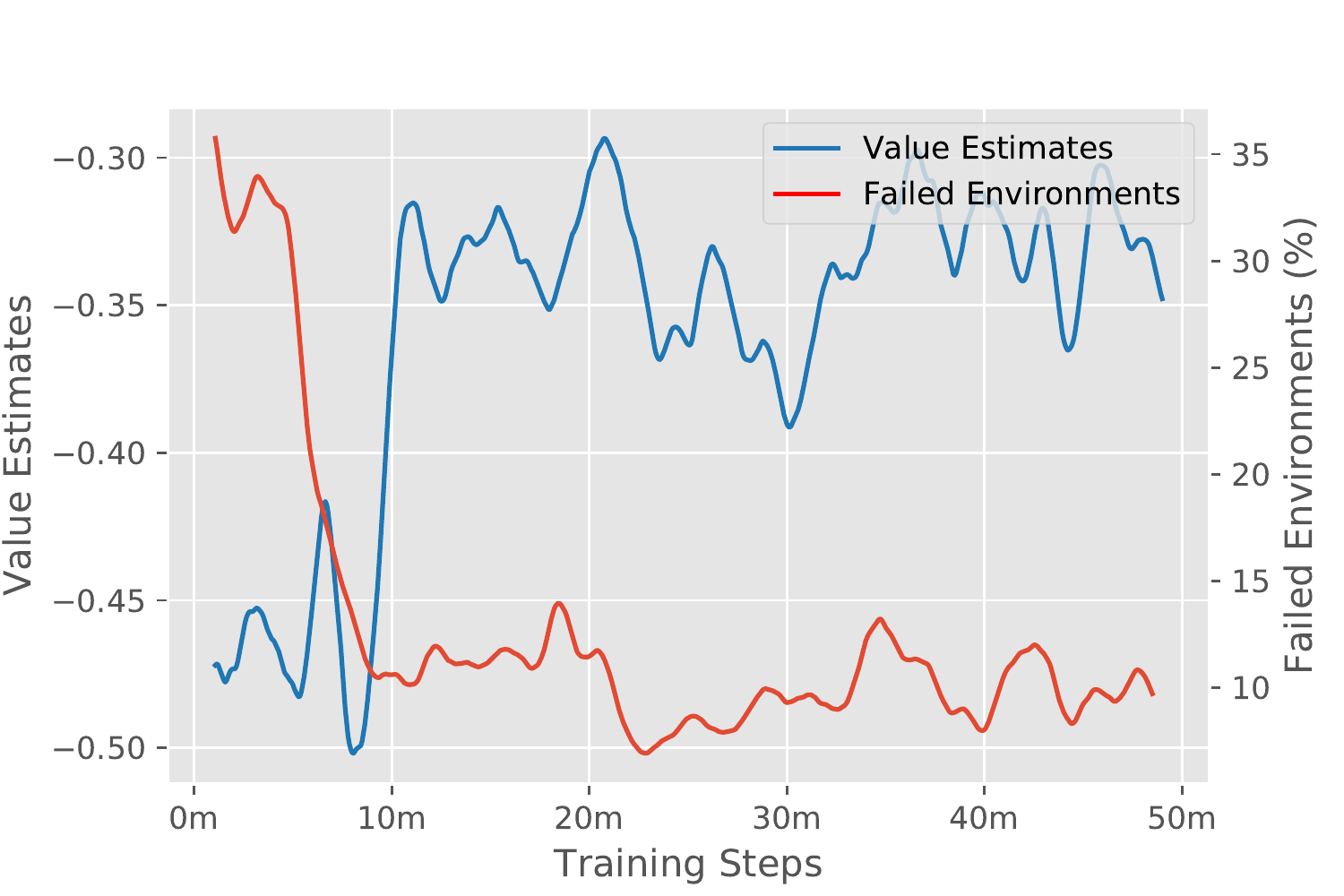}}
\caption{Percentage of failed, uncompilable, environments vs estimated value for the failed environments. As there are less failures, the agent does not accurately value these environments}
\label{fig:est_fail}
\end{center}
\end{figure}





\section{Conclusion}
\label{conclusion}

Generative Playing Networks is a novel framework for procedural content generation informed by the behavior and value estimates of a learning agent.
The method requires no data, nor domain knowledge, but is computationally intensive. The process is fully differentiable, allowing the agent to directly communicate with the generator about what designs it ``wants''. By playing levels it designs for itself it can learn a distribution of playable levels at a difficulty that match its skill. As reinforcement learning agents increase in performance, so will the complexity of the levels that GPN can discover.

In this paper, we have introduced an algorithm with a reward system based around complexity and an RL update rule that allows for efficient value estimates. The framework can also allow for richer and more interesting reward functions based around the agent's interactions with the environment.



\section*{Software and Data}

The Codebase for all these experiments can be found here:
\begin{center}
\url{https://github.com/pbontrager/GenerativePlayingNetworks}
\end{center}

\section*{Acknowledgements}
We would like to acknowledge Ahmed Khalifa for helpful discussions.

\bibliography{bibliography}

\begin{thebibliography}{10}
\providecommand{\url}[1]{#1}
\csname url@samestyle\endcsname
\providecommand{\newblock}{\relax}
\providecommand{\bibinfo}[2]{#2}
\providecommand{\BIBentrySTDinterwordspacing}{\spaceskip=0pt\relax}
\providecommand{\BIBentryALTinterwordstretchfactor}{4}
\providecommand{\BIBentryALTinterwordspacing}{\spaceskip=\fontdimen2\font plus
\BIBentryALTinterwordstretchfactor\fontdimen3\font minus
  \fontdimen4\font\relax}
\providecommand{\BIBforeignlanguage}[2]{{%
\expandafter\ifx\csname l@#1\endcsname\relax
\typeout{** WARNING: IEEEtran.bst: No hyphenation pattern has been}%
\typeout{** loaded for the language `#1'. Using the pattern for}%
\typeout{** the default language instead.}%
\else
\language=\csname l@#1\endcsname
\fi
#2}}
\providecommand{\BIBdecl}{\relax}
\BIBdecl

\bibitem{shaker2016procedural}
N.~Shaker, J.~Togelius, and M.~J. Nelson, \emph{Procedural content generation
  in games}.\hskip 1em plus 0.5em minus 0.4em\relax Springer, 2016.

\bibitem{risi2019procedural}
S.~Risi and J.~Togelius, ``Procedural content generation: From automatically
  generating game levels to increasing generality in machine learning,''
  \emph{arXiv preprint arXiv:1911.13071}, 2019.

\bibitem{cobbe2018quantifying}
K.~Cobbe, O.~Klimov, C.~Hesse, T.~Kim, and J.~Schulman, ``Quantifying
  generalization in reinforcement learning,'' \emph{arXiv preprint
  arXiv:1812.02341}, 2018.

\bibitem{zhang2018study}
C.~Zhang, O.~Vinyals, R.~Munos, and S.~Bengio, ``A study on overfitting in deep
  reinforcement learning,'' \emph{arXiv preprint arXiv:1804.06893}, 2018.

\bibitem{justesen2018illuminating}
N.~Justesen, R.~R. Torrado, P.~Bontrager, A.~Khalifa, J.~Togelius, and S.~Risi,
  ``Illuminating generalization in deep reinforcement learning through
  procedural level generation,'' in \emph{AAAI Workshop on Reinforcement
  Learning in Games}, 2019.

\bibitem{togelius2011search}
J.~Togelius, G.~N. Yannakakis, K.~O. Stanley, and C.~Browne, ``Search-based
  procedural content generation: A taxonomy and survey,'' \emph{IEEE
  Transactions on Computational Intelligence and AI in Games}, vol.~3, no.~3,
  pp. 172--186, 2011.

\bibitem{summerville2018procedural}
A.~Summerville, S.~Snodgrass, M.~Guzdial, C.~Holmg{\aa}rd, A.~K. Hoover,
  A.~Isaksen, A.~Nealen, and J.~Togelius, ``Procedural content generation via
  machine learning (pcgml),'' \emph{IEEE Transactions on Games}, vol.~10,
  no.~3, pp. 257--270, 2018.

\bibitem{bontrager2018deepmasterprints}
P.~Bontrager, A.~Roy, J.~Togelius, N.~Memon, and A.~Ross, ``Deepmasterprints:
  Generating masterprints for dictionary attacks via latent variable
  evolution,'' in \emph{2018 IEEE 9th International Conference on Biometrics
  Theory, Applications and Systems (BTAS)}.\hskip 1em plus 0.5em minus
  0.4em\relax IEEE, 2018, pp. 1--9.

\bibitem{volz2018evolving}
V.~Volz, J.~Schrum, J.~Liu, S.~M. Lucas, A.~Smith, and S.~Risi, ``Evolving
  mario levels in the latent space of a deep convolutional generative
  adversarial network,'' in \emph{Proceedings of the Genetic and Evolutionary
  Computation Conference}, 2018, pp. 221--228.

\bibitem{khalifa2020pcgrl}
A.~Khalifa, P.~Bontrager, S.~Earle, and J.~Togelius, ``Pcgrl: Procedural
  content generation via reinforcement learning,'' \emph{arXiv preprint
  arXiv:2001.09212}, 2020.

\bibitem{togelius2008experiment}
J.~Togelius and J.~Schmidhuber, ``An experiment in automatic game design,'' in
  \emph{2008 IEEE Symposium On Computational Intelligence and Games}.\hskip 1em
  plus 0.5em minus 0.4em\relax IEEE, 2008, pp. 111--118.

\bibitem{wang2019poet}
R.~Wang, J.~Lehman, J.~Clune, and K.~O. Stanley, ``Poet: open-ended coevolution
  of environments and their optimized solutions,'' in \emph{Proceedings of the
  Genetic and Evolutionary Computation Conference}, 2019, pp. 142--151.

\bibitem{dennis2020emergent}
M.~Dennis, N.~Jaques, E.~Vinitsky, A.~Bayen, S.~Russell, A.~Critch, and
  S.~Levine, ``Emergent complexity and zero-shot transfer via unsupervised
  environment design,'' \emph{arXiv preprint arXiv:2012.02096}, 2020.

\bibitem{gisslen2021adversarial}
L.~Gissl{\'e}n, A.~Eakins, C.~Gordillo, J.~Bergdahl, and K.~Tollmar,
  ``Adversarial reinforcement learning for procedural content generation,''
  \emph{arXiv preprint arXiv:2103.04847}, 2021.

\bibitem{goodfellow2014generative}
I.~Goodfellow, J.~Pouget-Abadie, M.~Mirza, B.~Xu, D.~Warde-Farley, S.~Ozair,
  A.~Courville, and Y.~Bengio, ``Generative adversarial nets,'' in
  \emph{Advances in neural information processing systems}, 2014, pp.
  2672--2680.

\bibitem{mnih2016asynchronous}
V.~Mnih, A.~P. Badia, M.~Mirza, A.~Graves, T.~Lillicrap, T.~Harley, D.~Silver,
  and K.~Kavukcuoglu, ``Asynchronous methods for deep reinforcement learning,''
  in \emph{International conference on machine learning}, 2016, pp. 1928--1937.

\bibitem{justesen2019deep}
N.~Justesen, P.~Bontrager, J.~Togelius, and S.~Risi, ``Deep learning for video
  game playing,'' \emph{IEEE Transactions on Games}, 2019.

\bibitem{sutton2018reinforcement}
R.~S. Sutton and A.~G. Barto, \emph{Reinforcement learning: An
  introduction}.\hskip 1em plus 0.5em minus 0.4em\relax MIT press, 2018.

\bibitem{nguyen2015deep}
A.~Nguyen, J.~Yosinski, and J.~Clune, ``Deep neural networks are easily fooled:
  High confidence predictions for unrecognizable images,'' in \emph{Proceedings
  of the IEEE conference on computer vision and pattern recognition}, 2015, pp.
  427--436.

\bibitem{nguyen2016synthesizing}
A.~Nguyen, A.~Dosovitskiy, J.~Yosinski, T.~Brox, and J.~Clune, ``Synthesizing
  the preferred inputs for neurons in neural networks via deep generator
  networks,'' in \emph{Advances in neural information processing systems},
  2016, pp. 3387--3395.

\bibitem{kingma2014auto}
D.~P. Kingma and M.~Welling, ``Auto-encoding variational bayes,'' \emph{stat},
  vol. 1050, p.~1, 2014.

\bibitem{perez2019general}
D.~Perez-Liebana, J.~Liu, A.~Khalifa, R.~D. Gaina, J.~Togelius, and S.~M.
  Lucas, ``General video game ai: A multitrack framework for evaluating agents,
  games, and content generation algorithms,'' \emph{IEEE Transactions on
  Games}, vol.~11, no.~3, pp. 195--214, 2019.

\bibitem{torrado2018deep}
R.~R. Torrado, P.~Bontrager, J.~Togelius, J.~Liu, and D.~Perez-Liebana, ``Deep
  reinforcement learning for general video game ai,'' in \emph{2018 IEEE
  Conference on Computational Intelligence and Games (CIG)}.\hskip 1em plus
  0.5em minus 0.4em\relax IEEE, 2018, pp. 1--8.

\bibitem{brockman2016openai}
G.~Brockman, V.~Cheung, L.~Pettersson, J.~Schneider, J.~Schulman, J.~Tang, and
  W.~Zaremba, ``Openai gym,'' \emph{arXiv preprint arXiv:1606.01540}, 2016.

\bibitem{radford2015unsupervised}
A.~Radford, L.~Metz, and S.~Chintala, ``Unsupervised representation learning
  with deep convolutional generative adversarial networks,'' \emph{arXiv
  preprint arXiv:1511.06434}, 2015.

\bibitem{shi2016real}
W.~Shi, J.~Caballero, F.~Husz{\'a}r, J.~Totz, A.~P. Aitken, R.~Bishop,
  D.~Rueckert, and Z.~Wang, ``Real-time single image and video super-resolution
  using an efficient sub-pixel convolutional neural network,'' in
  \emph{Proceedings of the IEEE conference on computer vision and pattern
  recognition}, 2016, pp. 1874--1883.

\bibitem{he2016deep}
K.~He, X.~Zhang, S.~Ren, and J.~Sun, ``Deep residual learning for image
  recognition,'' in \emph{Proceedings of the IEEE conference on computer vision
  and pattern recognition}, 2016, pp. 770--778.

\end{thebibliography}
\bibliographystyle{IEEEtran}

\end{document}